\pdfoutput=1

\documentclass[11pt]{article}

\usepackage[final]{acl}

\usepackage{times}
\usepackage{latexsym}
\usepackage{booktabs}
\usepackage[T1]{fontenc}

\usepackage[utf8]{inputenc}

\usepackage{microtype}

\usepackage{inconsolata}

\usepackage{graphicx}

\usepackage{multirow}
\usepackage{xspace}
\usepackage[textsize=scriptsize]{todonotes}

\newenvironment{icompact}{
  \begin{list}{$\bullet$}{
    \itemindent -.05em
    \parsep 0pt plus 1pt
    \partopsep 0pt plus 1pt
    \topsep 2pt plus 2pt minus 2pt
    \itemsep 0pt plus 1.3pt
    \parskip 0pt plus 2pt
    \leftmargin 0.13in}
      }
{\normalsize
\end{list}
}

\usepackage{textcomp}
\usepackage{verbatim}
\usepackage{listings}
\usepackage{xcolor}
\usepackage{listings}
\usepackage{url}
\usepackage{xcolor}
\newcommand{\sys}{\textsc{RippleCOT}\xspace}
\newcommand{\para}[1]{\vspace{2pt}\noindent{\textbf{#1}\xspace\vspace{0.1pt}}}

\title{\textsc{RippleCOT}: Amplifying Ripple Effect of Knowledge Editing in Language Models via Chain-of-Thought In-Context Learning}

\author{
  \textbf{Zihao Zhao\textsuperscript{1}},
  \textbf{Yuchen Yang\textsuperscript{1}},
  \textbf{Yijiang Li\textsuperscript{2}},
  \textbf{Yinzhi Cao\textsuperscript{1}}
\\
  \textsuperscript{1}Johns Hopkins University \\
  \textsuperscript{2}University of California San Diego\\
  \texttt{\{zzhao71, yc.yang, yinzhi.cao\}@jhu.edu, yijiangli@ucsd.edu}
}

\begin{document}
\maketitle

\begin{abstract}
The ripple effect poses a significant challenge in knowledge editing for large language models. Namely, when a single fact is edited, the model struggles to accurately update the related facts in a sequence, which is evaluated by multi-hop questions linked to a chain of related facts. Recent strategies have moved away from traditional parameter updates to more flexible, less computation-intensive methods, proven to be more effective in the ripple effect. In-context learning (ICL) editing uses a simple demonstration \texttt{Imagine that + new fact} to guide LLMs, but struggles with complex multi-hop questions as the new fact alone fails to specify the chain of facts involved in such scenarios. Besides, memory-based editing maintains additional storage for all edits and related facts, requiring continuous updates to stay effective. As a result of the design limitations, the challenge remains, with the highest accuracy being only 33.8\% on the \textsc{MQuAKE-cf} benchmarks for Vicuna-7B. To address this, we propose \sys, a novel ICL editing approach integrating Chain-of-Thought (COT) reasoning. \sys structures demonstrations as \{\texttt new fact, \texttt question, \texttt thought, \texttt answer\}, incorporating a \emph{thought} component to identify and decompose the multi-hop logic within questions. This approach effectively guides the model through complex multi-hop questions with chains of related facts.
Comprehensive experiments demonstrate that \sys significantly outperforms the state-of-the-art on the ripple effect, achieving accuracy gains ranging from 7.8\% to 87.1\%. \sys is open-source and available at \url{https://github.com/zzhao71/RippleCOT}

\end{abstract}
\section{Introduction}

\begin{figure}[t!]
  \centering
  \scriptsize
  \includegraphics[width=1\linewidth]{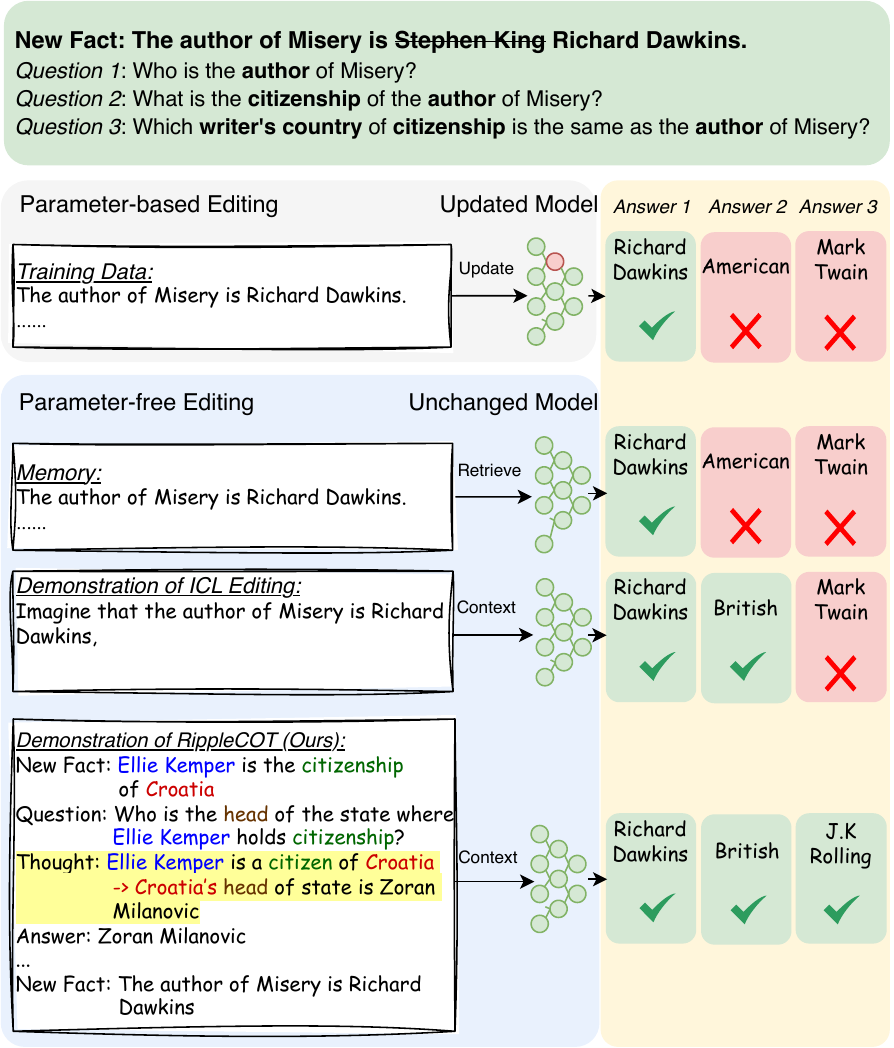}
  \caption{An illustration of \sys and existing parameter-based and parameter-free knowledge editing methods addressing the ripple effect via multi-hop questions. }\label{fig:intro}
\end{figure}

As large language models (LLMs) become more prevalent in various sectors, their limitations, such as storing inaccurate or sensitive knowledge, pose growing concerns~\cite{Dhingra_2022, carlini2021extracting,wolf2019huggingface}. This has led to the development of knowledge editing methods aimed at updating the facts. The ripple effect represents a significant challenge in knowledge editing for LLMs that was not explored until very recently~\cite{cohen2023evaluating}. When one fact is edited in a model, the ripple effect refers to the chain of related facts that should be updated following the edited one, which is evaluated by the multi-hop questions~\cite{zhong2023mquake} linked to a chain of facts. Figure~\ref{fig:intro} illustrates an example of the multi-hop question: if we modify the author of Misery to Richard Dawkins, the related multi-hop facts, such as the citizenship of the author of Misery, should also be updated.

Conventional parameter-based editing methods, such as fine-tuning~\cite{zhu2020modifying} or matrix computation~\cite{meng2022mass}, update the model's parameters to recall specific edited facts effectively but risk catastrophic forgetting~\cite{zheng2023can} and were proven failure on the ripple effect~\cite{cohen2023evaluating}. The edits are limited to the facts within the training data and struggle with related but untrained facts. Recent parameter-free editing approaches, like memory-based editing~\cite{mitchell2022memory}, also face challenges when related facts fall outside the maintained memory's scope. In-context learning (ICL) editing uses the simple prompt \texttt{Imagine that + new fact} to help the model recall the new fact. However, it struggles with complex, multi-hop questions because the new fact alone does not specify the chain of facts within such scenarios. Due to those limitations, the best result~\cite{zhong2023mquake} achieves only a 33.8\% accuracy on the \textsc{MQuAKE-cf} benchmarks with Vicuna-7B model, with other methods performing around 15\%.

To address this, we propose to integrate COT reasoning into the ICL framework, guiding LLMs to process multi-hop questions sequentially. While we observe the direct use of a "Think step by step" COT prompt improving performance, it falls short in open-source models with limited reasoning capacities. To arrive at a better solution, we develop \sys, by structuring the demonstration to $\texttt{(new\hspace{0.25em}fact}$, $\texttt{question}$, $\texttt{thought}$, $\texttt{answer)}$. \sys operates in two stages: demonstration generation and refinement. During generation, \sys identifies multiple relationships and missing items within the defined fact triplets $(s, r, o)$~\cite{cohen2023evaluating}, i.e., subject, relation, and object. For example, in Figure~\ref{fig:intro}, the question involves multiple relations: $\texttt{citizenship}$ and $\texttt{head}$. \sys decomposes the questions into $(\texttt{Ellie Kemper}$, $\texttt{citizenship}$, $?)$ and $(?$, $\texttt{head}$, $?^*)$. Given the new fact $(\texttt{Ellie Kemper}$, $\texttt{citizenship}$, $\texttt{Croatia})$, the $?$ is identified as $\texttt{Croatia}$. The $\texttt{thought}$ then becomes \emph{Ellie Kemper is a citizen of Croatia $\rightarrow$ Croatia’s head of state is Zoran Milanovic}. This effectively triggers the COT reasoning ability of the LLMs, significantly improving the ripple effect for more complex questions. In the refinement stage, \sys selects the top-k candidates among generated $(\texttt{new\ fact, question, thought, answer})$ pairs whose questions have the highest cosine similarity with the task question. 

Beyond COT reasoning ability, \sys offers several advantages. First, \sys operates without altering model parameters, resulting in lower computational costs and enabling efficient adaptation to multi-hop questions for a single edit. It also supports multiple editing scenarios, which have been less explored by the literature, such as accurately updating the related facts if changing the President of the United States from Obama to Trump and then to Biden. Second, our COT demonstration generation is automatic and highly flexible, tailored to task-specific questions. We have designed and evaluated multiple methods for generating demonstrations, including human selection from benchmark datasets along with the existing approaches, few-shot generation using GPT-4o ~\cite{achiam2023gpt} based on selected references, and zero-shot generation with GPT-4o, identifying the optimal approach for enhancing ripple effects. Additionally, \sys can be integrated with existing knowledge editing methods to further improve ripple effect performance.

In summary, this paper has three main contributions:
\begin{icompact}
\item We propose a novel ICL knowledge editing framework with automatic COT demonstration generation and refinement, namely \sys.
\item We explore different ways for generating COT demonstration, namely full-shot selection, few-shot generation, and zero-shot generation
\item \sys significantly improves accuracy, ranging from 7.8\% to 87.1\%, in addressing ripple effects on multi-hop questions, as demonstrated on the \textsc{RippleEdit}~\citep{cohen2023evaluating} and \textsc{MQuAKE}~\citep{zhong2023mquake} datasets.
\end{icompact}

\section{Related Work}
\subsection{Knowledge Editing}
Knowledge editing methods include parameter-based methods~\cite{mitchell2021fast, meng2022locating, dong2022calibrating} and parameter-free methods~\cite{zheng2023can, zhong2023mquake, Wang2024DeepEditKE, chen2024robust}. In parameter-based editing methods, Fintuning~\cite{zhu2020modifying} uses gradient descent to update model parameters based on the edit; MEND~\cite{mitchell2021fast} introduces hyper-networks that convert the gradients to model parameter changes; ROME~\cite{meng2022locating} introduces causal tracing that locates and updates the parameters responsible for factual associations. However, these methods can result in catastrophic forgetting of previously learned knowledge, perform poorly at generalizing edits to related facts, and are computationally expensive.
Parameter-free knowledge editing methods, mainly ICL editing~\cite{zheng2023can}, utilize demonstration to prompt the model to generate outputs aligned with the injected knowledge. Later in \citep{cohen2023evaluating}, simply prompting the model with the edited fact without demonstration can achieve better performance than parameter-based methods. Combined with retrieval-augmented methods, recent parameter-free methods such as EREN~\cite{chen2024robust} and MeLLo~\cite{zhong2023mquake} achieve SOTA in multiple metrics\citep{chen2024robust, zhong2023mquake}.

\subsection{Ripple Effect}
A common under-addressed problem of all knowledge editing methods is the propagation of knowledge updates to other logically connected facts, which is referred to as the ripple effect. Cohen et al.~\cite{cohen2023evaluating} first define and categorize the ripple effects into logical generalization, compositionality I \& II, subject aliasing, preservation, and relation specificity, each denoting a different logical pattern. Compositionality I \& II involves two-hop questions, in which the models perform the worst. This observation is recapitulated by Zhong et al.~\cite{zhong2023mquake}, where the performance of the edited model is evaluated by 2,3,4-hop questions. Overall, the knowledge editing performance decreases as the intermediate logical steps increase.

\section{Problem Formulation}

Knowledge editing aims to update the fact triplet from $(s, r, o)$ to $(s, r, o^*)$, where $s$ is the subject, $r$ the relation, $o$ the original object and $o^*$ the new object. These triplets are formulated~\cite{petroni_18} as prompt templates $p(s,r, \emptyset)$—for example, with $s=\texttt{Stephen King}$, $r=\texttt{Citizenship}$ and $\emptyset$ is a place holder for the object, the prompt is: \emph{The citizenship of Stephen King is \textunderscore}. Using a language model $f: \mathcal{X} \to \mathcal{Y}$, which processes input prompt $x \in \mathcal{X}$ to generate output $y \in \mathcal{Y}$, we probe the model with $p(s,r, \emptyset)$. The output $f(p(s,r, \emptyset))$ should match the original object $o$, such as \emph{American}. After editing, the model $f^*$ should return $f^*(p(s,r, \emptyset))$, matching the updated object $o^*$, such as \emph{British}.

\subsection{In-Context Knowledge Editing}

Given a new fact triplet $(s, r, o^*)$, In-context knowledge editing injects it via an input prompt starting with the prefix \emph{``Imagine that"}~\cite{cohen2023evaluating}, following the template $p(s,r,o^*)$ We denoted this new fact injection as $e(s,r,o^*) =$ $\texttt{``Imagine that"}$ $+$ $p(s,r,o^*)$. The edited model is then defined as $f* = f \circ (e(s,r,o^*))$. To verify the edit, we query the edited model with $f^*(p(s,r, \emptyset))$ and check if it successfully recalls $o^*$.

\subsection{Ripple Effect with Multi-hop Questions}

The ripple effect is assessed through multi-hop questions~\cite{zhong2023mquake}. Imagine a chain of facts $\mathcal{Q} = \{ (s_1, r_1, o_1), \ldots, (s_n, r_n, o_n) \}$, where each object $o_i$ serves as the subject $s_{i+1}$ in the subsequent fact. We refer to the set of relations as $\mathcal{R} = \{r_1, \ldots, r_n\}$ and the set of subjects as $\mathcal{S} = \{s_1, \ldots, s_n\}$. The multi-hop questions are formulated using $\mathcal{Q}$ that begins with the head entity $s_1$ till the $r_n$, and the answer is the tail entity $o_n$. For instance, consider the question \emph{What is the citizenship of the author of Misery?} composed of the fact chains $\{\texttt{Misery}, \texttt{Author}, \texttt{Stephen King}\}$ and $\{$\texttt{Stephen King}, \texttt{Citizen}, \texttt{American}$\}$. If we update the first fact to $\{\texttt{Misery}, \texttt{Author}, \texttt{Ellie Kemper}\}$, the edited model $f^*$ is validated by checking if the response for the above question is \texttt{British} instead of \texttt{American}, reflecting the related fact $\{\texttt{Ellie Kemper}, \texttt{Citizen}, \texttt{British}\}$.

\section{\sys}

As discussed above, knowledge editing faces the challenge of ripple effects where a sequence of related facts should also be updated to arrive to the correct answer to some particular question. This chain of related facts for knowledge editing resembles a chain of thoughts in reasoning which motivates us to integrate CoT reasoning into the ICL pipeline and propose \sys as a unified solution. 

\subsection{\sys Formulation}
\label{sec:formulation}
For each knowledge editing, we construct $k$ demonstrations $\mathcal{D} = \{d_1, \dots, d_k\}$, which $d_i\in \mathcal{D}$ consist of four main components: new fact, question, thought, and answer. 

\para{New facts.}$(s_1, r_1, o^*_1)$, which are examples of information designated for edits.

\para{Questions.}Multi-hop questions $p(\mathcal{S}, \mathcal{R}, \emptyset)$ with $s\in \mathcal{S}$ and $r\in \mathcal{R}$, which are formulated to probe the related facts following the new facts, thereby assessing the ripple effects of edits.

\para{Thoughts.} Break down the questions according to each relation $r_1, \ldots, r_n \in R$ as follows: 
\[
\left\{
\begin{array}{l}
\{s_1, r_1, o^*_1\} \\
\{s_2^* = o^*_1, r_2, o^*_2\} \\
\vdots \\
\{s_{n-1}^* = o^*_{n-1}, r_{n-1}, o^*_n\}
\end{array}
\right.
\]
\para{Answers.} $o^*_n$, which provides the exact answer to the question, derived from the logical reasoning in the thoughts section.

The goal of the demonstration is to allow the model to generate the correct answer through its COT reasoning ability, establishing clear connections between the new facts and their related facts to the final answers.

\subsection{Demonstration Generation}
To generate $k$ demonstrations $\mathcal{D} = \{d_1, \dots, d_k\}$ with the COT formulation, we explore three approaches:

\para{Full-shot Selection.} 
$\forall d_i\in \mathcal{D}$, the \texttt{\{new fact, question, thought, answer\}} set is randomly selected from the \textsc{MQuAKE} benchmark. This ensures high-quality, logically coherent contexts that serve as reliable examples for the model. 

\para{Few-shot Generation.} 
To generalize the demonstration beyond the scope of \textsc{MQuAKE},
\sys first creates a reference set $\mathcal{D}_{\mathtt{refer}}$ containing a few demonstrations using the human selection. This reference set is then used to guide LLMs, which have shown remarkable reasoning and instruction-following ability, in generating demonstrations with a similar format. We evaluate both GPT-4o and GPT-J generated demonstrations, the prompt is as follows:
\begin{lstlisting}
Your task is to genereate knowledge editing examples for in context learning.
You need to first generate the knowledge being edited (fact being changed) and then ask a question that requires multi-hop (multi-step) reasoning. Finally you need to provide a answer with step-by-step reasoning in concise format.

Example: $\mathcal{D}_{\mathtt{refer}}$

Please respond in the following format without any markdown.
New Fact: <knowledge being editted>
Question: <question that requires multi-step reasoning>
Thought: <step-by-step reasoning in concise format>
Answer: <answer with step-by-step reasoning in concise format>

Please generate {$k$} knowledge editing examples. Please respond only the generated examples in the above format without any markdown or additional text.
\end{lstlisting}

\para{Zero-shot Generation}
\sys explore the Zero-shot Generation ability ~\cite{ramesh2021zero} of LLMs to directly generate examples following specific formats. Specifically, we remove the reference set $\mathcal{D}_{\mathtt{refer}}$, using only the COT format introduced in Section \ref{sec:formulation}. We use both GPT-4o and GPT-J to generate the demonstrations using the above prompt, omitting the line \texttt{``Example: $\mathcal{D}_{\mathtt{refer}}$"}.

\subsection{Demonstration Refinement}
After demonstration generation, we refine the demonstration by ordering it by the similarity ~\cite{lu2021fantastically} between the \texttt{question} components in the $\mathcal{D}$, denoted as $\{q_{\mathtt{demo}}\}_{i=1}^k$, and the question that we want the model to answer, denoted by $\{q_{\mathtt{target}}\}_{i=1}^k$. We follow Liu et al.~\cite{liu2021makes} to use the all-MiniLM-L6-v2~\cite{wang2020minilm} to get embeddings  ~\cite{reimers2019sentence} denoted as $\{\mathcal{E}(q_{\mathtt{demo}})\}_{i=1}^k$ and $\{\mathcal{E}(q_{\mathtt{target}})\}_{i=1}^k$ respectively. The similarity $\{m\}_i^k$ is calculated by the cosine similarity ~\cite{huang2008similarity} with each $m_i$:
\begin{equation}
\scriptstyle
     m_i = \frac{\mathcal{E}(q_{\mathtt{demo}})_i \cdot \mathcal{E}(q_{\mathtt{target}})_i}{\sqrt{||\mathcal{E}(q_{\mathtt{demo}})_i||_2  \cdot ||\mathcal{E}(q_{\mathtt{target}})_i||_2}}
\end{equation}

 Then, \sys select the top-$t$ demonstrations from $\{q_{\mathtt{demo}}\}_{i=1}^k$ with with the highest $\{m\}_i^k$. This approach ensures that the most relevant demonstrations are selected, thereby improving the overall performance of the model.

\section{Experiments}

\subsection{Experiment Setup}
We primarily assess our method using the \textsc{MQuAKE}~\citep{zhong2023mquake}  and  \textsc{RippleEdit}~\cite{cohen2023evaluating} dataset with the models GPT-J (6B) ~\cite{gpt-j}, Vicuna-7B ~\cite{zheng2024judging}, and GPT-3 ~\cite{brown2020language, ouyang2022training}. We adopt the accuracy metric from previous work~\cite{zhong2023mquake, cohen2023evaluating}, where an answer is deemed correct if the model's output contains the expected answer. We set our \emph{default} setting as the full-shot selection with $k=5$ demonstrations. 

\subsubsection{Dataset}

\para{\textsc{RippleEdit}.}This dataset contains counterfactual knowledge editing examples ~\cite{meng2022locating}. It is divided into three different subsets. The popular subset contains edits on popular entities in wiki data ~\cite{vrandevcic2014wikidata}; the random subset contains random entities; the recent subset contains recently added entities. We primarily focus on the popular subset.

\para{\textsc{MQuAKE}.}This dataset is used to test the edited model's multi-hop question-answering ability, which contains 2,3,4-hop questions.

\subsubsection{Baseline}

We conduct a comparative analysis of \sys against several established techniques: Fine-tuning (FT)~\cite{zhu2020modifying}, MEND~\cite{mitchell2021fast}, ROME~\cite{meng2022mass}, DeepEdit~\cite{Wang2024DeepEditKE}, IKE~\cite{cohen2023evaluating} and MEMIT~\cite{meng2022locating}, as well as our proposed BaseCOT, which adds a ``Think step by step" prompt ~\cite{kojima2022large} after the question, as the baseline method for the \sys approach.

\para{FT:}FT employs gradient descent to update model parameters based on the edits, directly modifying the weights to reflect the new information.

\para{MEND:}MEND trains a hypernetwork to transform raw fine-tuning gradients based on an edited fact, creating targeted weight updates to integrate new factual content.

\para{ROME:}ROME identifies and localizes factual knowledge within specific Transformer layers, then updates the feedforward networks in those layers to incorporate new facts.

\para{MeLLo:}MeLLo stores edited facts externally. During runtime, related facts are retrieved, and conflict detection ensures appropriate edited outputs.

\para{MEMIT:}MEMIT extends ROME by enabling simultaneous editing of a large set of facts. It updates feedforward networks across multiple layers, effectively encoding a broader range of factual information.

\para{DeepEdit:}This method views knowledge editing as a constrained decoding problem, ensuring outputs meet the proposed semantic constraints. DeepEdit uses a depth-first search-based progressive decoding technique for efficient updates without retraining.

\para{IKE:} The ICL editing approach with a demonstration as \emph{``Imagine that" + new fact}.

\subsection{Comparison with Baselines}

\begin{table*}[t!]
\scriptsize
\renewcommand{\arraystretch}{1.5}
\setlength{\tabcolsep}{10pt} 
  \centering
  \begin{tabular}{l|c|cccc|cccc} 
  \hline
    \toprule
    \multirow{2}{*}{\textbf{Model}} & \multirow{2}{*}{\textbf{Data}} & \multicolumn{4}{c|}{\textbf{Parameter-based Methods}} & \multicolumn{4}{c}{\textbf{Parameter-free Methods}} \\
    \cline{3-10}
    & & MEND & FT & ROME & MENIT & MELLO & IKE & BaseCOT & \bf \sys \\
    \midrule
    \multirow{3}{*}{\textbf{GPT-J}} & \textbf{\textsc{\textbf{MQuAKE-cf}}} & 11.5 & 1.9  & 18.1 & 12.3 & 41.2 & 67.2 & 59.3 & \textbf{81.7} \\
    & \textbf{\textsc{\textbf{MQuAKE-t}}} & 38.2 & 0.2 & 11.3 & 4.8 & 46.8 & 71.8 & 66.9 & \textbf{83.2} \\
    & \textbf{\textsc{\textbf{RippleEdit}}} & - & - & 48.4 & 50.2 & - & 26.1 & 30.2 & \textbf{47.5}\\
    \midrule
    \multirow{3}{*}{\textbf{Vicuna-7B}} & \textbf{\textsc{\textbf{MQuAKE-cf}}}  &  8.4 & 0.2 & 12.2 & 9.0 & 33.8 &  20.6 & 20.0 &  \textbf{87.3}\\
    & \textbf{\textsc{\textbf{MQuAKE-t}}} & 33.9 & 8.2 & 9.3 & 5.8 & 51.3 & 30.7 & 39.7 & \textbf{78.9}\\
    & \textbf{\textsc{\textbf{RippleEdit}}} & - & - & 68.7 & 59.3 & - & 82.0 & 43.2 & \textbf{89.8}\\
    
    \bottomrule
  \end{tabular}
  \caption{\label{tab: ep-1}Comparison between existing knowledge editing methods with \sys and in \textsc{MQuAKE} and \textsc{RippleEdit} dataset. The number represents the accuracy (\%) in answering the questions after the model is edited.}
\end{table*}

To mimic the human-written chain-of-thought context, we extract new facts, questions, thoughts, and answers from multi-hop questions in the MQuAKE dataset. Each question in this dataset is a 2-hop, 3-hop, or 4-hop question. We combine each subquestion and its corresponding answer into a single sentence to form the thought process. Following the approach of Zhong et al.~\citep{zhong2023mquake}, we post three similar questions. If one of these questions yields an accurate answer, we consider the model to have successfully edited the new facts. The default number of contexts used is five. The results are presented in Table~\ref{tab: ep-1}.

Our model shows much better performance compared to previously proposed models in all three datasets. Furthermore, compared to BaseCOT, our method still shows better performance, which reinforces that our method helps improve the model's reasoning ability and amplifies the ripple effects.
\subsection{Performance on One-time Edit}

\para{The number of edited instances.}
Following the methodology of \cite{zhong2023mquake}, we split the dataset into groups of $g$ instances, where $g$ values are 1, 100, 1000, and 3000. For a higher number of edited facts, \sys introduces a dynamic retrieval method based on similarity measures between the thought and the stored new facts. The evaluation prompt becomes:
\begin{verbatim}
[5-shot demonstrations]
[New facts: m facts line by line 
retrieved from the given 3000 facts]
[Question]
\end{verbatim}
The $m$ new facts are selected based on their similarity to the generated thoughts.

The details of the dynamic retrieval process are as follows. Note that $m$ is not fixed, because a single question may relate to multiple edited facts. For example, for the question, “What is the capital of the country to which Lou Pearlman belonged?”, the relevant facts might be “Lou Pearlman is a citizen of India” and “The capital of India is Taloga.” To address this, \sys retrieves up to $m$ rounds with one fact per round and employs an early stopping criterion if no contradiction is detected during a self-check. The setting of self-check follows \cite{zhong2023mquake} and \cite{Wang2024DeepEditKE} for identifying contradictions between the retrieved facts and the answer. For each retrieval round, \sys selects one fact from the remaining stored new facts that are most similar to the generated thought, and append it to the [New facts] prompt. This novel dynamic retrieval mechanism is also a key contribution of \sys in enhancing the retrieval process. The results are presented in Table~\ref{tab: ep-2-1}.

Parameter-based methods face a large decline when the edit instances increase because it is very hard to update parameters for editing numerous instances. Additionally, retrieval accuracy becomes low when the number of edited instances increases, causing the accuracy of related edited questions to drop. However, our method focuses on teaching the model to think with the provided logic, so our method does not decline when the number of edited instances increases, demonstrating its potential for handling a large number of edits.

\begin{table}[t!]
\scriptsize
\renewcommand{\arraystretch}{1.5}
\setlength{\tabcolsep}{7pt} 
  \centering
  \begin{tabular}{ll|cccc} 
  \toprule
    \multirow{2}{*}{\textbf{Model}} & \multirow{2}{*}{\textbf{Method}} & \multicolumn{4}{c}{\textbf{\# Edit Instances}}  \\ 
    & & 1 & 100 & 1000 & 3000 \\
  \midrule
  \multirow{4}{*}{\textbf{GPT-J}}
  & MEMIT &  12.3 & 9.8 & 8.1 & 1.8\\
  & MEND &  11.5 & 9.1 & 4.3 & 3.5\\
  & MeLLo & 20.3 & 12.5 & 10.4 & 9.8\\
  & \textbf{\sys} & \textbf{80.1} &\textbf{79.3} & \textbf{78.1} & \textbf{79.9}\\
  \midrule
  \multirow{2}{*}{\textbf{Vicuna-7B}}
  & MeLLo & 20.3 & 11.9 &11.0 &10.2 \\
  & \textbf{\sys} & \textbf{87.3} & \textbf{82.8} & \textbf{80.8} & \textbf{85.7}\\
    \midrule
  \multirow{2}{*}{\textbf{GPT-3}}
  & MeLLo & 68.7 & 50.5 & 43.6& 41.2\\
  & \textbf{\sys}& \textbf{89.7} & \textbf{88.6} & \textbf{82.7} & \textbf{80.9}\\
    \bottomrule
  \end{tabular}
  \caption{\label{tab: ep-2-1} Performance of \sys and baselines with GPT-J, Vicuna-7B. We evaluate the number of edited instances once as ${1, 100, 1000, 3000}$ on \textsc{MQuaKE-cf}. We include the best results reported by the baselines for comparison.}
\end{table}

\para{The number of hops.}
The dataset contains examples for 2-hop, 3-hop, and 4-hop questions. For multi-hop questions, we adhere to the previous prompt standard and concatenate these sentences to form the thoughts. This method assesses the model's ability to apply learned facts. As the number of hops increases, the model must utilize all learned new facts and apply logical reasoning to generalize the question. Table~\ref{tab: ep-2-2} demonstrates that typically, as the number of hops increases from 2 to 4, accuracy decreases. However, in \sys, this decline is minimized compared to other methods, indicating our method's superior capability in enabling the model to apply new facts effectively.
\begin{table}[t!]
\scriptsize
\renewcommand{\arraystretch}{1.5}
\setlength{\tabcolsep}{11pt} 
  \centering
  \begin{tabular}{l|cccc} 
  \toprule
  \textbf{Method} & \bf 2-hop & \textbf{3-hop} & \textbf{4-hop} & \textbf{All} \\
  \midrule
  FT & 3.7 & 1.4 & 0.5 & 1.9 \\
  MEND & 13.9 & 11.3& 9.5& 11.5\\
  ROME & 33.8 & 9.1 & 11.4 & 18.1\\
  MEMIT& 22.5 &6.0 &8.4 &12.3\\
  MeLLo & 47.5 & 27.2 & 45.3 & 42.1 \\
  \textbf{\sys} & \textbf{80.2}& \textbf{78.8} & \textbf{79.2} & \textbf{80.1}\\ 
    \bottomrule
  \end{tabular}
  \caption{ Performance of \sys and baselines with GPT-J. We evaluate the number of hop-in questions from \textsc{MQuaKE-cf} as ${2, 3, 4}$, and all are referred to as "All".  We include the best results reported by the baselines for comparison. }
  \label{tab: ep-2-2}
\end{table}

\subsection{Performance in medical applications}
We conducted experiments on the MedCF dataset~\cite{xu2024editing}, a benchmark for medical question-answer tasks. We follow Xu et al.~\cite{xu2024editing} to evaluate the Meditron-7B model. The Table~\ref{medical} shows the applicability of \sys to knowledge editing in the medical domain. Unlike BaseCOT which relies heavily on the model's reasoning ability, \sys tailors the thought process for knowledge editing, effectively decomposing multi-hop logic in questions.
\begin{table}
  \centering
  \scriptsize
\renewcommand{\arraystretch}{1.5}
\setlength{\tabcolsep}{18pt} 
  \begin{tabular}{l|cc} 
    \toprule
     & \textbf{BaseCOT} & \textbf{\sys}  \\
    \midrule
    Meditron (7B) & 65.3 & 99.9  \\
    \bottomrule
  \end{tabular}
  \caption{We compare \sys with BaseCOT on the MedCF dataset using the Meditron-7B model.}
    \label{medical}
\end{table}

\subsection{Performance on Multi-time Edit}
Previously, to our knowledge, all methods have evaluated knowledge editing using multi-hop questions or simple question-answering ~\cite{wang2022modern} under one-time editing. However, in real life, it is common to update knowledge multiple times. For instance, in the context of presidential elections, the president of the United States changes every 4 or 8 years, necessitating repeated updates to this knowledge. Due to the lack of datasets evaluating this aspect, we decided to modify the answers twice. For example, as shown in Figure~\ref{fig:intro}, we change the author of Misery from Stephen King to Richard Dawkins, and then to a new author, Ernest Hemingway. We applied similar changes to a total of randomly selected 200 datasets by altering the answers to multi-hop questions to mimic multiple edits in real life.

Table~\ref{tab: ab-7} shows the number of edits significantly impacts retrieval-based methods, which struggle with these kinds of problems. When conflicting knowledge is injected into the knowledge base, retrieval accuracy decreases, leading to lower accuracy for multi-hop questions. However, our method, \sys, is less affected by multiple edits. The performance of \sys remains relatively stable even with multiple edits, provided the demonstrations do not change.

\begin{table}
  \label{mquake}
  \centering
  \scriptsize
\renewcommand{\arraystretch}{1.5}
\setlength{\tabcolsep}{2pt} 
  \begin{tabular}{l|cccc} 
    \toprule
     & \textbf{Mello} & \textbf{Mello after M.E.} & \textbf{\sys} & \textbf{\sys after M.E.}  \\
    \midrule
    GPT-J & 41.2 & 8.9 & 81.7 & 79.9 \\
    \bottomrule
  \end{tabular}
  \caption{We compare the performance across different numbers of demonstrations using the GPT-J and Vicuna models, evaluated on the \textsc{MQuaKE-cf} dataset.  M.E. stands for Multiple Editing.}
  \label{tab: ab-7}
\end{table}

\subsection{Combination with Baselines}
\begin{figure}[t]
  \includegraphics[width=\columnwidth]{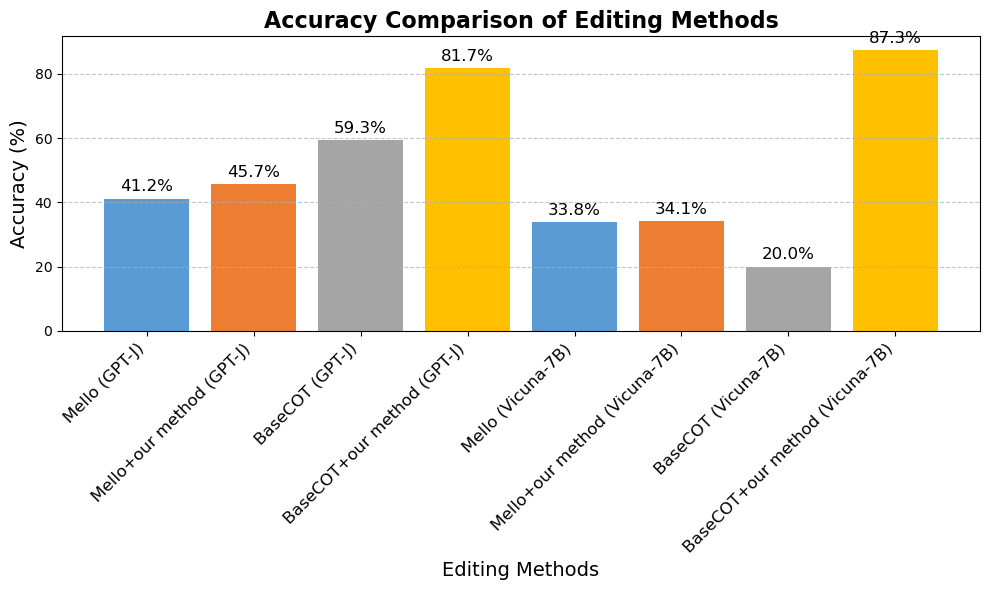}
  \caption{Comparative analysis of the performance enhancements achieved by our proposed method when applied to two baseline models, Mello and BaseCOT, in both the GPT-J and Vicuna-7B architectures.}
  \label{fig:experiments}
\end{figure}

In this study, we integrate our proposed method into existing approaches and baselines to evaluate its effectiveness in improving performance. The performance enhancements observed, as illustrated in Figure~\ref{fig:experiments}
, demonstrate the significant impact of our method in amplifying ripple effects.

\subsection{Ablation Study}

In this section, we ablate on different components of \sys, which are generation strategy, number of referenced demonstrations used for the few-shot generation, and number of demonstrations selected for the full-shot selection.

\subsubsection{Demonstration Generation}

\para{Generation Strategy.} Table~\ref{tab: ab-1} indicates that few-shot generation may significantly amplify ripple effects. The comparable performance between few-shot generation and human selection, particularly with GPT-4o's few-shot performance surpassing that of human selection (human-written chain-of-thought content), suggests a viable alternative for automatic context generation. We utilize GPT-4o and GPT-J (6B) models to generate demonstrations using both few-shot and zero-shot generation ~\cite{levy2017zero}. For zero-shot learning, we input the prompt directly, allowing the model to autonomously generate the result. For few-shot learning, we extract several contexts from the \textsc{MQuaKE} dataset, which includes new facts, questions, thoughts, and answers, and prompt the model to generate the context in a similar format. The generated content is then used as context for \sys. We impose several criteria on the generated context: it must comprehensively include all four sections (facts, questions, thoughts, and answers), and the answer should be concise, ideally a term or a few words. The similarity between GPT-J and GPT-4o generation suggests that performance is optimal when provided with several high-quality examples. However, in the zero-shot scenario, GPT-4o may still outperform due to its superior capability in generating reasonable chain-of-thought prompts.

\begin{table}[h]
  \label{mquake}
  \centering
  \scriptsize
\renewcommand{\arraystretch}{1.5}
\setlength{\tabcolsep}{1.6pt} 
  \begin{tabular}{l|ccc} 
    \toprule
    & Zero-shot Generation & Few-shot Generation& Full-shot Selection \\
    \midrule
    GPT-J & 69.5 & 80.8 & \multirow{4}{*}{\bf 81.7} \\
    ChatGPT-3.5 & 74.2 & 81.2 \\
    ChatGPT-4 & 80.8 & 81.9 \\
    GPT-4o & 72.7 & 82.3\\
    \bottomrule
  \end{tabular}
  \caption{Comparison between different demonstration generation. We use the demonstration generated by GPT-J and GPT-4o for zero-shot and few-shot generation. The result is on \textsc{MQuaKE-cf}  dataset. }
  \label{tab: ab-1}
\end{table}

\para{How many referenced demonstrations are needed for the few-shot generation?} We try to answer this question in Table \ref{tab:num_human_demon}. By alternating the different number of references given to the LLMs to few-shot generate demonstrations, we observe that with more references, performance steadily grows to the best. We also want to emphasize that a competitive performance can be achieved using as little as 1 reference, which illustrates that \sys can work well even with little reference selected from $\textsc{MQuaKE}$ and thus can be generalized to other datasets.
\begin{table}[h]
  \label{mquake}
  \centering
  \scriptsize

\renewcommand{\arraystretch}{1.5}
\setlength{\tabcolsep}{10pt} 
  \begin{tabular}{l|ccccc} 
    \toprule
    \# References & 1 & 2& 3 & 4 & 5 \\
    \midrule
    ChatGPT-3.5 & 68.9 & 80.2 & 78.6 & 79.3 & 81.2 \\
    ChatGPT-4 & 78.7 & 81.0 & 81.0 & 78.0 & 81.9 \\
    GPT4-o & 80.8 & 67.9 & 70.5 & 71.0 & 82.3 \\
    \bottomrule
  \end{tabular}
  \caption{We compare the performance of the few-shot demonstration given different numbers of human reference examples. The result is on \textsc{MQuaKE-cf} dataset. }
  \label{tab:num_human_demon}
\end{table}

\para{How many demonstrations are needed to tackle the ripple effect in knowledge editing?} We answer this question in Table \ref{tab: ab-9}. We experimented with using 1, 2, 5, 10, and 20 demonstrations to evaluate the performance. As the number of contexts increases, computational complexity also rises. Our goal is to identify an optimal number of demonstrations that elicit high-quality model outputs. As shown in Table~\ref{tab: ab-9}, using 5 demonstrations appears to be an effective choice. While using 20 demonstrations offers a slight improvement in accuracy, considering the trade-off between accuracy gains and the cost of time and computational resources, 5 demonstrations represent a practical and efficient choice.

\begin{table}[h]
  \label{mquake}
  \centering
  \scriptsize
\renewcommand{\arraystretch}{1.5}
\setlength{\tabcolsep}{8pt} 
  \begin{tabular}{l|ccccc} 
    \toprule
    \# Demonstrations & 1 & 2& 5 & 10 & 20 \\
    \midrule
    GPT-J & 69.5 & 75.8 & 81.7 & 80.9 &  82.0\\
    Vicuna-7B & 51.9 &  62.8 & 87.3& 85.8 & 88.4\\
    \bottomrule
  \end{tabular}
  \caption{compare the performance across different numbers of demonstrations using the GPT-J and Vicuna-7B models, evaluated on the \textsc{MQuaKE-cf} dataset. }
  \label{tab: ab-9}
\end{table}

\subsubsection{Demonstration Refinement}
As discussed in our methodology, we employ cosine similarity to select demonstrations. During the generation process, we initially generate 20 candidate contexts and then select 5 of these based on cosine similarity. This approach results in approximately a 5\% performance increase compared to random selection, underscoring the importance of incorporating cosine similarity into our method.

\section{Conclusion}
\sys has demonstrated superior performance relative to existing approaches. Through our experiments, we have highlighted the importance of our method in amplifying the ripple effect, as well as its flexibility in integration with other existing methods. Additionally, our analysis of chain-of-thought generation provides valuable insights for automatic generation. By combining the inherent flexibility and improved efficacy of in-context editing, our method can significantly streamline the process of knowledge updating, facilitating more accurate and contextually relevant model responses across various domains.

\section*{Limitations}
Our study, while promising, has several notable limitations that should be addressed in future work:

\begin{icompact}
    \item \textbf{Limited Dataset Scope.} There are limited benchmarks for analyzing ripple effects, especially for multiple edits. We conducted experiments on only two datasets. We hope that, in the future, a larger dataset will be developed, encompassing various scenarios such as questions related to several parallel facts, to enable a more comprehensive evaluation.

    \item \textbf{Assumption of LLM Capabilities.} Our approach assumes that the employed LLMs possess sufficient capabilities to handle knowledge editing and chain-of-thought (CoT) reasoning. However, if sub-optimal LLMs are used, the effectiveness of the proposed methods may be compromised, leading to diminished overall performance.

    \item \textbf{Bias in Edits.} The creation of multiple edits to simulate real-life scenarios may inadvertently introduce biases. These biases might not accurately reflect the complexity and variability of natural knowledge updates. It is crucial to develop more objective and systematic methods for generating edits to ensure the authenticity and relevance of the scenarios used in experiments.


\end{icompact}

Addressing these limitations will be vital for advancing the field of knowledge editing and improving the effectiveness and reliability of methods like \sys in real-world applications.

\section*{Potential Negative Social Impact}
Our commitment to ethical research practices guided our methodology and implementation throughout the study, however, \sys may raise the following negative impacts:

 Firstly, we acknowledge the importance of ensuring the accuracy and integrity of information in language models. The ability to edit knowledge within these models must be approached with caution to prevent the propagation of misinformation. This means that our approach may be maliciously employed to distort, manipulate, or propagate misinformation. We raise this potential negative social impact here to highlight the need for stringent safeguards and monitoring mechanisms. Researchers and practitioners utilizing \sys must implement robust verification processes to ensure that only accurate and verified information is introduced into language models. 

 Secondly, we are aware of the potential biases that may be introduced through manual edits and the limitations of the datasets used. This means that using LLMs or the limited current datasets on knowledge editing might inherit the biases in LLMs or the current datasets.

By addressing these potential negative social impacts, we aim to contribute to the responsible advancement of knowledge editing technologies, ensuring they are used to enhance the reliability and effectiveness of language models in various applications.

\section*{Acknowledgments}

We would like to thank the anonymous reviewers for their helpful comments and
feedback.
This work was supported in part by Johns Hopkins University Institute for
Assured Autonomy (IAA) with grants 80052272 and 80052273,  National
Science Foundation (NSF) under grants CNS-21-31859, CNS-21-12562, CNS-19-37786, CNS-19-37787, and CNS-18-54000, as well as Army Research Office (ARO) under grant No. W911NF2110182. The views and conclusions contained herein are those of the authors
and should not be interpreted as necessarily representing the official policies or
endorsements, either expressed or implied, of NSF, ARO, or JHU-IAA.

\bibliography{custom}
\appendix
\section{Appendix}
\subsection{Example Prompts and Generation Results}

\begin{lstlisting}
{Demonstrations}
New Fact: Lou Pearlman is a citizen of India, The capital of India is Taloga
Question: What is the capital of the country to which Lou Pearlman belonged?

New fact: the author of Misery is Richard Dawkins.  
Question: What is the nationality of the author of Misery.  
Thought: The author of Misery is Richard Dawkins. Richard Dawkins is a citizen of United Kingdom. Therefore, the nationality of the author of Misery is British.  
Answer: British  

New fact: The capital of United States of America is El Campu.  
Question: What is the capital city of the country that Michael Feinstein is a citizen of ?  
Thought: Michael Feinstein is a citizen of United States of America. The capital of United States of America is El Campu. Thus, the capital city of the country that Michael Feinstein is a citizen of is El Campu.  
Answer: El Campu  

Thought: Lou Pearlman is a citizen of India. The capital of the country of which Lou Pearlman is a citizen is Taloga.
Answer: Taloga
\end{lstlisting}
\subsection{Comparison of Different CoT Methods}
Our evaluation in Table~\ref{various_cot} also shows that \sys outperforms the standard CoT approach, Base-COT with "think step by step," as well as other advanced CoT methods, i.e., Self-generated-COT \cite{wang2023self} that prompts the model to split a complex question into several sub-questions, while Least-to-most-COT \cite{zhou2022least} let the model generate and arrange the sub-questions from easy to hard. We also compare more advanced prompting such as Self-consistency \cite{wang2022self} by generating several candidates and performing majority voting. RippleCOT also outperforms plain Self-consistency, demonstrating the significance of CoT in knowledge editing. Self-consistency essentially is self-ensembling which be combined with RippleCOT. We demonstrate that RippleCOT can be further boosted by self-consistency, as shown in the Table~\ref{various_cot}.

The results indicate that RippleCOT is well-customized for knowledge editing. This is because RippleCOT creates thoughts that break down questions based on relationships, i.e., key components in knowledge editing, helping the model learn how to solve multi-hop questions in knowledge editing. In contrast, other methods let the model determine how to divide the questions. If the model makes an error during the early stages of problem decomposition, it can affect the following steps and lead to an incorrect final answer.
\begin{table}
  \centering
  \scriptsize
\renewcommand{\arraystretch}{1.5}
\setlength{\tabcolsep}{0.5pt} 
  \begin{tabular}{l|cccccc} 
    \toprule
     Method  & \textbf{SG-COT} & \textbf{LtM-COT} & \textbf{SC} & \textbf{\sys} & \textbf{\sys+SC} \\
    \midrule
    GPT-J  & 54.1 & 74.9 & 34.2 & 87.3 & 90.1 \\
    \bottomrule
  \end{tabular}
  \caption{Comparison of different Chain-of-Thought (CoT) methods. SG-COT: Self-Generated CoT, LtM-COT: Least-to-Most CoT, SC: Self-Consistency, \sys: Your system.}
  \label{various_cot}
\end{table}

\subsection{Safety Evaluation}
We conducted a jailbreak attack \cite{huang2023catastrophic} before and after applying RippleCOT, and found that the attack success rate is unchanged (92\%) for the MaliciousInstruct dataset with the Vicuna-7B model under their setting w/o sys. prompt. This is because RippleCOT does not alter any model parameters during editing, thus it does not affect the model's safety level.

\subsection{More baseline with large models}
We added three large models, GPT-4-0125 (1.8T), GPT-4o and Claude-3.5 sonnet (while the exact parameter size isn't specified, it is the latest high-performing large model), to the table below, in addition to GPT-3 (175B), which have already been included in our paper. We observe that RippleCOT performs well on both smaller and larger models, whereas BaseCOT, which relies solely on the model's reasoning ability, is effective only for larger models with enhanced reasoning capabilities. The results is shown in Table~\ref{performance_comparison}.

\begin{table}
  \centering
  \scriptsize
\renewcommand{\arraystretch}{1.5}
\setlength{\tabcolsep}{3pt} 
  \begin{tabular}{l|cccc} 
    \toprule
     Method & \textbf{GPT-3 (175B)} & \textbf{Claude-3.5 sonnet} & \textbf{GPT-4o} & \textbf{GPT-4-0125} \\
    \midrule
    BaseCoT     & 72.3 & 75.8 & 79.1 & 79.1 \\
    RippleCoT   & 89.7 & 85.2 & 87.9 & 88.6 \\
    \bottomrule
  \end{tabular}
  \caption{Performance comparison between BaseCoT and RippleCoT across different models.}
  \label{performance_comparison}
\end{table}

\end{document}